\documentclass[twocolumn,pre,aps,showpacs,groupedaddress,nofootinbib]{revtex4-1}
\usepackage{graphicx}
\usepackage{amssymb}
\usepackage{textcomp}
\usepackage{amsmath}
\usepackage[normalem]{ulem} 

\usepackage{color,soul}
\usepackage{url}

\def\be{\begin{equation}}
\def\ee{\end{equation}}
\def\bea{\begin{eqnarray}}
\def\eea{\end{eqnarray}}
\def\bdm{\begin{displaymath}}
\def\edm{\end{displaymath}}
\def\ba{\begin{array}}
\def\ea{\end{array}}

\begin{document}

\title{Standing Wave Decomposition Gaussian Process}

\author{Chi-Ken Lu}
\author{Scott Cheng-Hsin Yang}
\author{Patrick Shafto}
\address{Department of Mathematics and Computer Science, Rutgers University, Newark, New Jersey 07302}

\date{\today}

\begin{abstract}

We propose a Standing Wave Decomposition (SWD) approximation to Gaussian Process regression (GP). 
GP involves a costly matrix inversion operation, which limits applicability to large data analysis. 
For an input space that can be approximated by a grid and when correlations among data are short-ranged,
the kernel matrix inversion can be replaced by analytic diagonalization using the SWD. 
We show that this approach applies to uni- and multi-dimensional input data, extends to include longer-range correlations, 
and the grid can be in a latent space and used as inducing points. 
Through simulations, we show that our approximate method applied to the squared exponential kernel outperforms existing 
methods in predictive accuracy per unit time in the regime where data are plentiful. Our SWD-GP is recommended for regression analyses where there is a relatively large amount of data and/or there are constraints on computation time.

\end{abstract}


\maketitle

\section{Introduction}

Gaussian Process (GP) regression~\cite{rasmussen2006gaussian} is powerful as well as elegant because of its ability to transform correlation into prediction. This nonparametric and Bayesian approach needs not specify a predetermined function that the data points fit to, and only assumes that any subset of data follows a joint Gaussian distribution characterized by the mean and covariance functions. GP is equivalent to a class of neural networks which have infinite number of hidden units~\cite{neal1997monte,williams1997computing}. As a flexible model for highly complicated functions and physical processes, GP is a familiar tool for physicists in many context, such as predicting melting temperature~\cite{seko2014machine}, inter-atomic potential~\cite{bartok2010gaussian}, fractional Brownian motion~\cite{sadhu2018generalized}, DNA replication kinetics~\cite{baker2012inferring,baker2014inferring}, gravitational waveform~\cite{doctor2017statistical}, etc. Furthermore, in the machine learning community, GP can be used for classification~\cite{rasmussen2006gaussian,morales2017remote}, data dimension reduction, and signal reconstruction~\cite{titsias2010bayesian}.

However, the scalability of GP regression is limited by its $O(N^3)$ training time for $N$ training data points. Inverting the kernel matrix is the critical step which essentially transforms correlation into prediction. The task of speeding up the GP training algorithm has been taken up by computer scientists who use inducing points and Nystr\"{o}m method to approximate the kernel matrix~\cite{williams2001using}, and, recently, by physicists~\cite{das2018continuous,zhao2015quantum} who employ quantum algorithms capable of solving system of linear equations~\cite{harrow2009quantum}. In this paper, we propose an approximating GP based on sparsifying the kernel matrix of a latent grid in the input space, which allows us to analytically diagonalize the kernel matrix by employing standing wave decomposition. Observing that correlation among nearest neighbors is most relevant to prediction, we approximate the full matrix with its tridiagonal matrix.\footnote{In a tridiagonal matrix, the only non-zero values are on the main diagonal, superdiagonal, and subdiagonal.} This tridiagonal matrix has analytic eigenvalues and eigenvectors resembling standing waves, a fact that has been extensively exploited in the studies of electronic structure of solids~\cite{marcus1993tight}. If the training data is on a grid, like time-series data, then the approximated kernel matrix can be immediately used in the usual GP prediction algorithm. When the given data is not on a grid in the input space, a latent grid is treated as inducing points and the function values are obtained by Bayesian projection from off-grid data. Our approach works best for the squared exponential (SE) kernel matrix, a very popular kernel for many machine learning tasks, because the off-diagonal matrix elements decay rapidly. For high dimensional grid data, the SE kernel matrix has a Kronecker product form, which makes the present approach applicable as well.     
 
One constraint of the approximated GP is that the ratio of length scale for SE kernel to the grid spacing, $\ell/\Delta$, has an upper bound because the kernel matrix must be semi-definite positive, which is guaranteed if the smallest eigenvalue is positive. The tridiagonal approximation works well when the neglected off-diagonal matrix elements have little effect on those small eigenvalues and the corresponding eigenvectors. An improvement to the tridiagonal approximation and the upper bound for $\ell/\Delta$ is the pentadiagonal approximation, which has the next nearest correlation included in the kernel matrix. The pentadiagonal matrix, however, does not have simple form of eigenvalues and eigenvectors. Thus, using the eigenvector of tridiagonal kernel matrix, we reconstruct the kernel matrix so that the next-nearest neighbor correlation terms are included and the eigenvectors are still exact with eigenvalues modified. In the end, a simulation with synthetic data is carried out with comparison with other GP approximations to highlight the condition under which the proposed method is at its best.  


The rest of the paper is structured as follows. 
In Sec.~\ref{GP_Overview}, an overview of GP regression along with an analytic example in terms of eigen basis decomposition is given. In Sec.~\ref{SW_Basis}, we introduce the approximate tridiagonal kernel matrix for a one-dimensional grid input and derive the standing wave eigen basis. The extension to two-dimensional grid input using the Kronecker product is provided in Sec.~\ref{multidimensional case}. To account for longer-ranged correlations, we demonstrate in Sec.~\ref{sec:recons} the reconstruction of a nearly pentadiagonal kernel matrix which allows the standing wave decomposition Gaussian process (SWD-GP) to be applied with longer length scale. Based on the above techniques, in Sec.~\ref{sec:latent} our SWD-GP is then combined with a latent grid (LG) to model data that do not do not lie on a grid. Our LGSWD-GP method is compared with popular existing methods, and the simulated accuracy and run time are reported in Sec.~\ref{sec:simu}. Sec.~\ref{sec:time} is devoted to the investigation of the time complexity of our approximated methods and a comparison with some popular algorithms. In the end, the related works and a brief discussion are given in Sec.~\ref{sec:relatedW} and Sec.~\ref{sec:diss}, respectively.

\section{Overview of GP}\label{GP_Overview}

Here we give an overview of GP by considering two unobserved values $y_1$ and $y_2$ at two distinct locations $x_1$ and $x_2$, respectively. To make it a tractable problem, we may assume the values follow a joint normal distribution, namely $p(y_1,y_2) = \mathcal N(0,\Sigma)$ with zero mean and convariance matrix 
\begin{equation}
 \Sigma = \sigma^2\left(\begin{array}{cc}
    1 & c \\
    c & 1
\end{array}\right)\:.   
\end{equation} The parameter $c$ is a function of distance $|x_1-x_2|$, which quantifies the correlation between $y_1$ and $y_2$. Here, $|c|\leq 1$ so that $\Sigma$ is positive semidefinite. It is then interesting to see how the joint distribution adjusts itself after one of the variable, say, $y_2$ has been observed. The way the correlation is transformed into prediction is seen by first decomposing the inverse of kernel matrix using the eigenvalues and eigenvectors, which yields $\Sigma^{-1} = \frac{1}{\lambda_+}\rm v_+v_+^T+\frac{1}{\lambda_-}\rm v_-v_-^T$. The eigenvalues are $\lambda_{\pm}=\sigma^2(1\pm c)$ and the corresponding eigenvectors read
\begin{equation}
    \rm v_{\pm} = \frac{1}{\sqrt{2}}
    \left(\begin{array}{cc}
         1  \\
         \pm 1 
    \end{array}\right)\:.
\end{equation} After some algebra, we find that the joint distribution is factorized into,
\begin{equation}\label{simpleGP}
    p(y_1,y_2)  \propto\exp{\left[-\frac{(y_1-cy_2)^2}{2(1-c^2)\sigma^2}\right]}\exp(-\frac{y_2^2}{2\sigma^2})\:. 
\end{equation} Thus, the adjusted distribution for $y_1$ becomes conditional on the observed $y_2$. It is easy to verify that $p(y_1|y_2) = p(y_1,y_2)/p(y_2)$ is still Gaussian, and the updated mean and variance $\mu_{1|2} = cy_2$ and $\sigma^2_{1|2} = (1-c^2)\sigma^2$, respectively.

Generalizing the above procedure to a multivariate joint distribution over the set of observed and unobserved variables $\{y_1,y_2,\cdots,y_N,y_*\}$ with a general kernel matrix $K$ constitutes the GP regression, which is to seek the underlying latent function $y=f(\bf x)$ mapped from the input points $\{{\bf x}_1, {\bf x}_2,\cdots,{\bf x}_N,{\bf x}_*\}$. Because of the Gaussianity, the correlation among the data leads to the conditional distribution over the unobserved variable, $p(y_*|y_{1:N})$, which is another Gaussian $\mathcal N(\mu_*,\sigma_*^2)$ with updated mean,
\begin{equation}\label{GP_mean}
    \mu_* = {\bf k}^T_* K^{-1}{\bf y}\:,
\end{equation} and variance,
\begin{equation}\label{GP_variance}
    \sigma_*^2 = \sigma^2 - {\bf k}^T_*K^{-1}{\bf k}_*\:, 
\end{equation} where $K$ denotes the kernel matrix associated with training data $\{({\bf x}_i,y_i)\}_{i=1}^N$ and ${\bf k}_*$ denotes the kernel matrix between training and test points $({\bf x}_*,y_*)$. Corresponding to the above example, $K=\sigma^2$, $k_*=c\sigma^2$, and $y=y_2$. 

In the following, we shall focus on the kernel matrix taken from the squared exponential kernel function $[K]_{lm}=k({\bf x}_l,{\bf x}_m)$,
\begin{equation}\label{K_matrix}
	k({\bf x,x'}) = \sigma^2\prod_{i=1}^d\exp\left[-\frac{(x_i-x'_i)^2}{2\ell^2}\right]\:,
\end{equation} where the product form appears due to the separable sum of squared distance along each dimension. The hyperparameters include the variance $\sigma^2$ and length scale $\ell$.

\section{Standing Wave Eigen Basis}\label{SW_Basis}


It is observed that the kernel matrix $K$ given by (\ref{K_matrix}) has a simplified form if all the input points $\{{\bf x}\}_{i=1}^M$ are on an one-dimensional grid since the matrix elements $k({\bf x}_i,{\bf x}_j)$ only depend on $|i-j|$, as in a Toeplitz matrix~\cite{gray2006toeplitz}. Here we make further simplification by retaining only the most relevant terms, namely the matrix elements $[K]_{ij}$ with $|i-j|\leq 1$, which is quite legitimate when the grid unit length $\Delta$ is larger than length scale. Notice that the rest of matrix elements decrease exponentially as $\alpha^{(i-j)^2}$ with $\alpha = \exp(-\frac{\Delta^2}{2\ell^2})$. The eigenvalue equations $K{\rm v} = \lambda{\rm v}$ in terms of the coefficients of eigenvector ${\rm v} = [c_1, c_2, ..., c_M]$ read
\begin{subequations}\label{eig_eq}
\begin{align}
	&\alpha c_2  + (1-\tilde\lambda)c_1 = 0\:, \label{first_eq}\\
	&\alpha c_{i+1} + (1-\tilde\lambda)c_i + \alpha c_{i-1} = 0\:,\label{mid_eq}\\
	&(1-\tilde\lambda)\ c_N + \alpha c_{N-1}= 0\:,\label{final_eq} 
\end{align}
\end{subequations} where the index $i = 2\cdots M-1$ applies in (\ref{mid_eq}). The eigenvalues are rescaled as $\tilde\lambda = \lambda/\sigma^2$. We observe that (\ref{eig_eq}) is similar with the energy Hamiltonian for an electron hopping on
semi-infinite one-dimensional tight-binding lattice~\cite{marcus1993tight}. Defining the z-transform associated with an auxiliary $\{c_i\}_{i=1}^{\infty}$ as $F(z) = \sum_{i=1}^{\infty}c_iz^{1-i}$, we find from (\ref{first_eq}) and (\ref{mid_eq}) that $F(z) = c_1[1-(\alpha+(1-\tilde\lambda)z)/(\alpha z^2+(1-\tilde\lambda)z+\alpha)]$. Moreover, the sequence $\{c_i\}$ can be reproduced from the contour integral ~\cite{marcus1993tight},
\begin{equation}
	c_i = c_1\left[\delta_{i1} - \frac{1}{2\pi i}\oint F(z) z^{i-2} dz \right]\:,
\end{equation} followed by employing the residue theorem. We note that the pair of poles of $F(z)$ are $z_{1,2} = e^{\pm i\theta}$ with which the eigenvalues $\tilde\lambda = 1+2\alpha\cos\theta$ can be consistent. Thus, the coefficients are shown to be,
\begin{equation}\label{solution}
	c_i = \frac{\sin i\theta}{\sin\theta} c_1\:,
\end{equation} for all nonnegative $i$. The final stage is to take care of (\ref{final_eq}), which is not necessarily consistent with the solution in (\ref{solution}) with arbitrary $\theta$ and $\tilde\lambda$. The only way to make (\ref{solution}) a valid solution is to set the auxiliary term $c_{M+1}=0$, which results in  
\begin{equation}\label{theta}
    \theta_k = k\frac{\pi}{M+1},\ \ \ k = 1, 2, 3,\cdots, M\:,
\end{equation} and the corresponding eigenvalues,
\begin{equation}\label{eigenvalue}
    \lambda_k = \sigma^2(1+2\alpha\cos\theta_k)\:.
\end{equation} 

\begin{figure}[ht]
\begin{center}
\centerline{\includegraphics[width=\columnwidth]{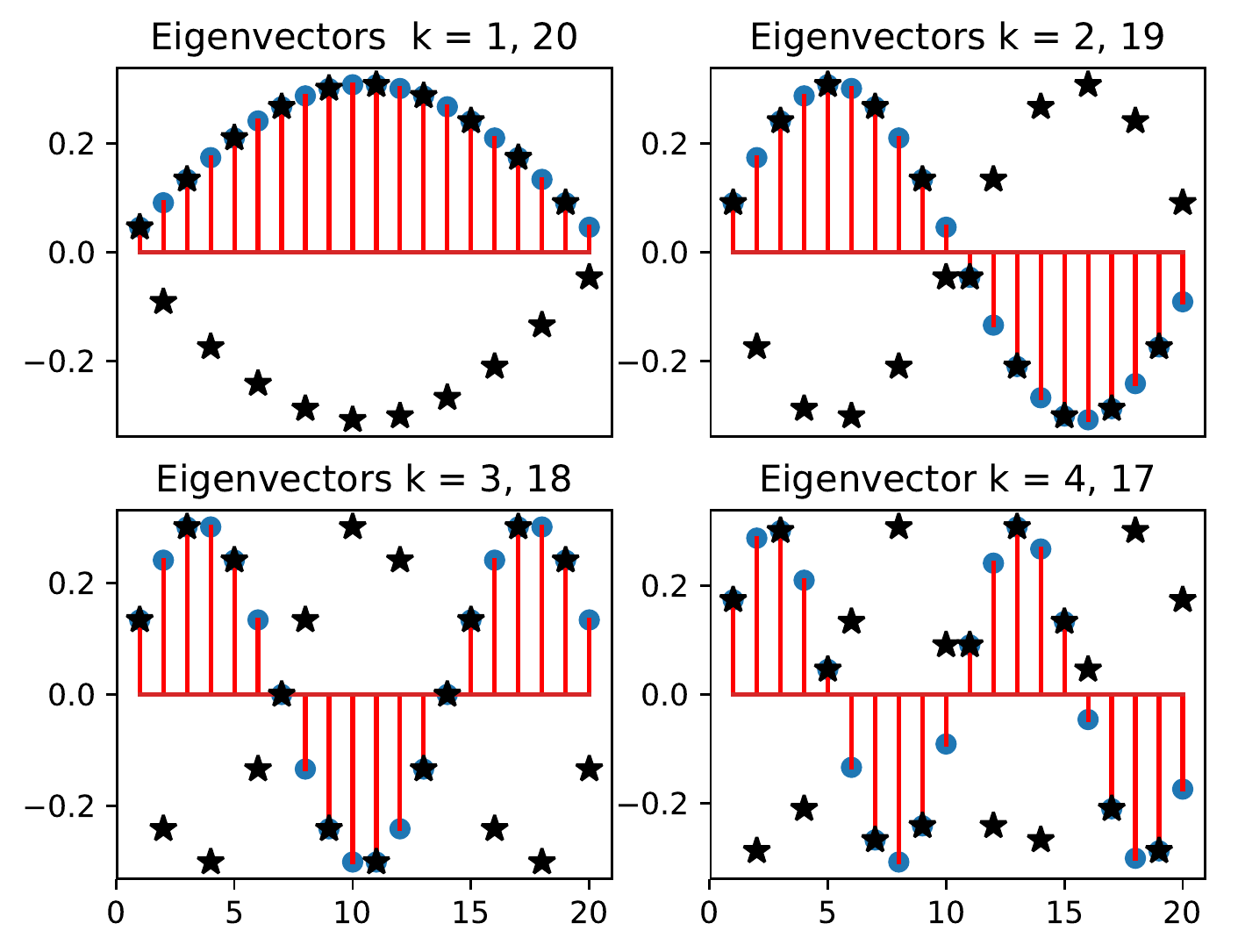}}
\caption{Four pairs of eigenvectors ${\rm v}_k$ and ${\rm v}_{M-k}$ are shown for the tridiagonal kernel matrix with $M$ = 20. Circle symbols denote the components of $[{\rm v}_k]$ while the star symbols represent its symmetric partner $[{\rm v}_{M-k+1}]$. Panels (a) to (d) correspond to $k$ = 1, 2, 3, and 4, respectively.}
\label{EigenVector}
\end{center}
\end{figure}

Consequently, we may conclude the orthonormal standing wave basis,
\begin{equation}\label{EigenBasis}
    [{\rm v}_k]_j = \frac{\sin j\theta_k}{\sqrt{\frac{M+1}{2}}}\:.
\end{equation} Fig.~\ref{EigenVector} demonstrates four pairs of symmetric eigenvectors $\{{\rm v}_k,{\rm v}_{M-k+1}\}$ of an 20$\times$20 kernel matrix $K$. As seen from the identity in (\ref{theta}), it is interesting to remark that the arguments $\theta_k$ and $\theta_{M-k+1}$ always sum up to $\pi$, which suggests the symmetric coefficients $[{\rm v}_k]_i = (-1)^{i+1}[{\rm v}_{M-k+1}]_i$. Such symmetry is also manifested in Fig.~\ref{EigenVector} where, for instance, in (a) the star symbols representing ${\rm v}_{20}$ alternate its sign whereas the dot symbols associating with ${\rm v}_{1}$ do not. Finally, the inverted kernel matrix can be simply decomposed as,
\begin{equation}\label{eq:swd_inverse}
	K^{-1} = \sum_{k=1}^{M} \frac{1}{\lambda_k}{\rm v}_k{\rm v}_k^{\rm T}\:.
\end{equation} 

\begin{figure}[h]
\begin{center}
\includegraphics[width=0.8\columnwidth]{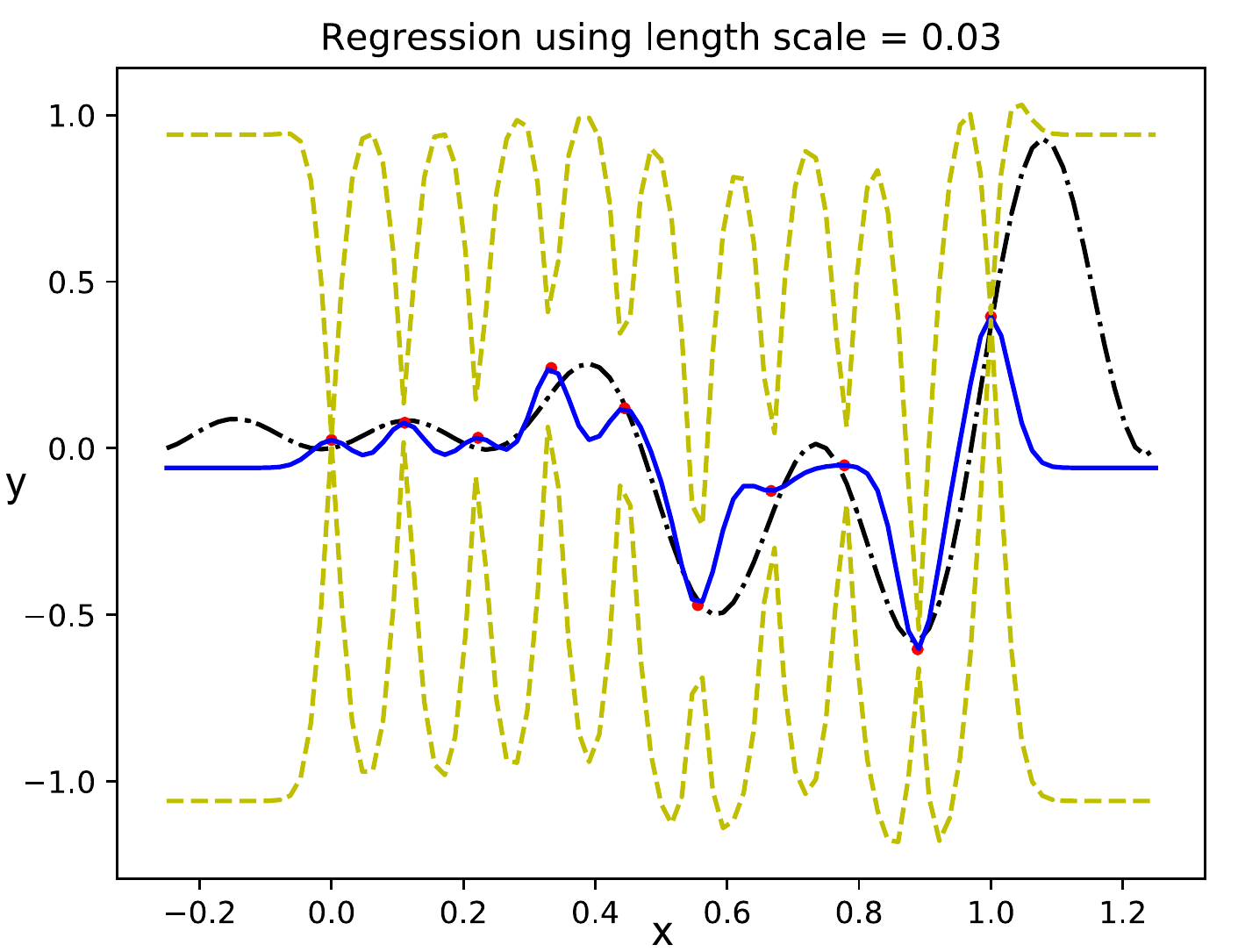}
\includegraphics[width=0.8\columnwidth]{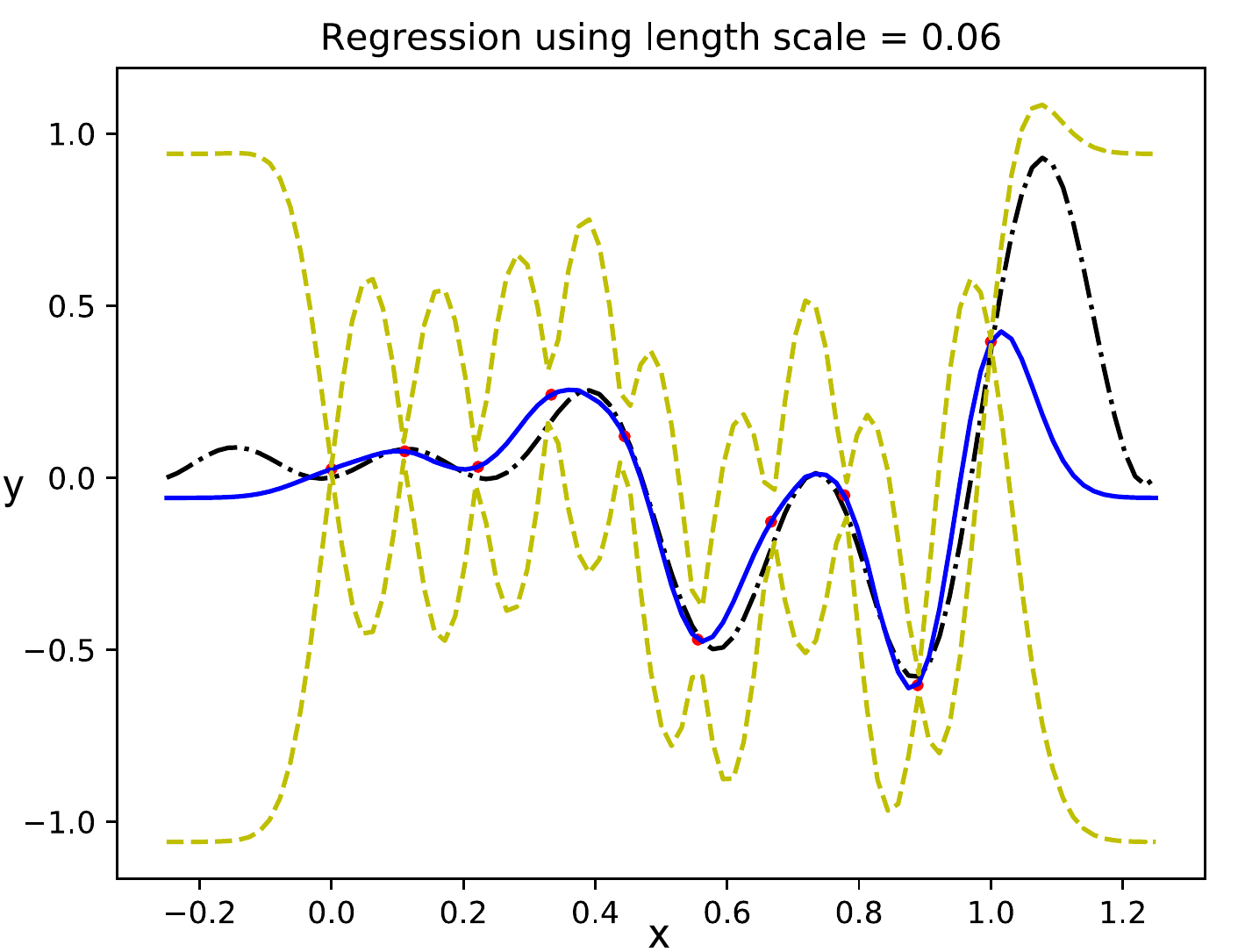}
\caption{Regression using on-grid data (red dots) generated by the function $f(x) = x\cos(2\pi x)\sin[4\pi(x+0.03)]$ (black dashed line). Blue line and yellow dashed lines represent the prediction mean $\mu_*$ of SWD-GP and plus/minus one variance $\sigma_*$ around the mean. The 10 grid points are uniformly placed in $[0,1]$. The length scale are $\ell = 0.03$ and 0.06, respectively, for (a) and (b).}
\label{Regression}
\end{center}
\end{figure}

Here, we note that the same sparse approximation which results in the tridiagonal $K$ must also be applied to the kernel matrix ${\bf k}_*$ so that the product ${\bf k}_*K^{-1}$ shall result in a row vector which is one at the $j$th element and zero otherwise when the test point coincides with the $j$th training point. The expression for $K^{-1}$ in (\ref{eq:swd_inverse}) suggests that the prediction mean of SWD-GP shall pass through the training points if all the eigenvalues are positive. Next, we employ the inverted kernel matrix (\ref{eq:swd_inverse}) in the GP learning algorithm (\ref{GP_mean}) and (\ref{GP_variance}), and apply this SWD-GP to the on-grid data, $\{x_i\}_{i=1}^{10}$, evenly spaced in $[0,1]$, and $\{y_i=f(x_i)\}$ with $f(x)=x \cos(2\pi x)\sin[24\pi(x+0.03)]$. Figure~\ref{Regression}(a) and (b) represent the results using $\ell/\Delta = 0.27$ and 0.54, respectively. The prediction mean associated with smaller $\ell/\Delta$ passes through all the training points, but is not able to capture the longer-distance features. Away from the training points, the variance $\sigma_*$ is close to the value of $\sigma$ used in (\ref{eigenvalue}), which is expected from the squared exponential kernel. On the other hand, the prediction mean and variance are optimal for the used length scale $\ell/\Delta=0.54$ because further increased $\ell/\Delta$ shall lead to zero or negative eigenvalues in (\ref{eigenvalue}).

\section{Multi-dimensional Grid}\label{multidimensional case}

\begin{figure}[h]
\centerline{\includegraphics[width=0.8\columnwidth]{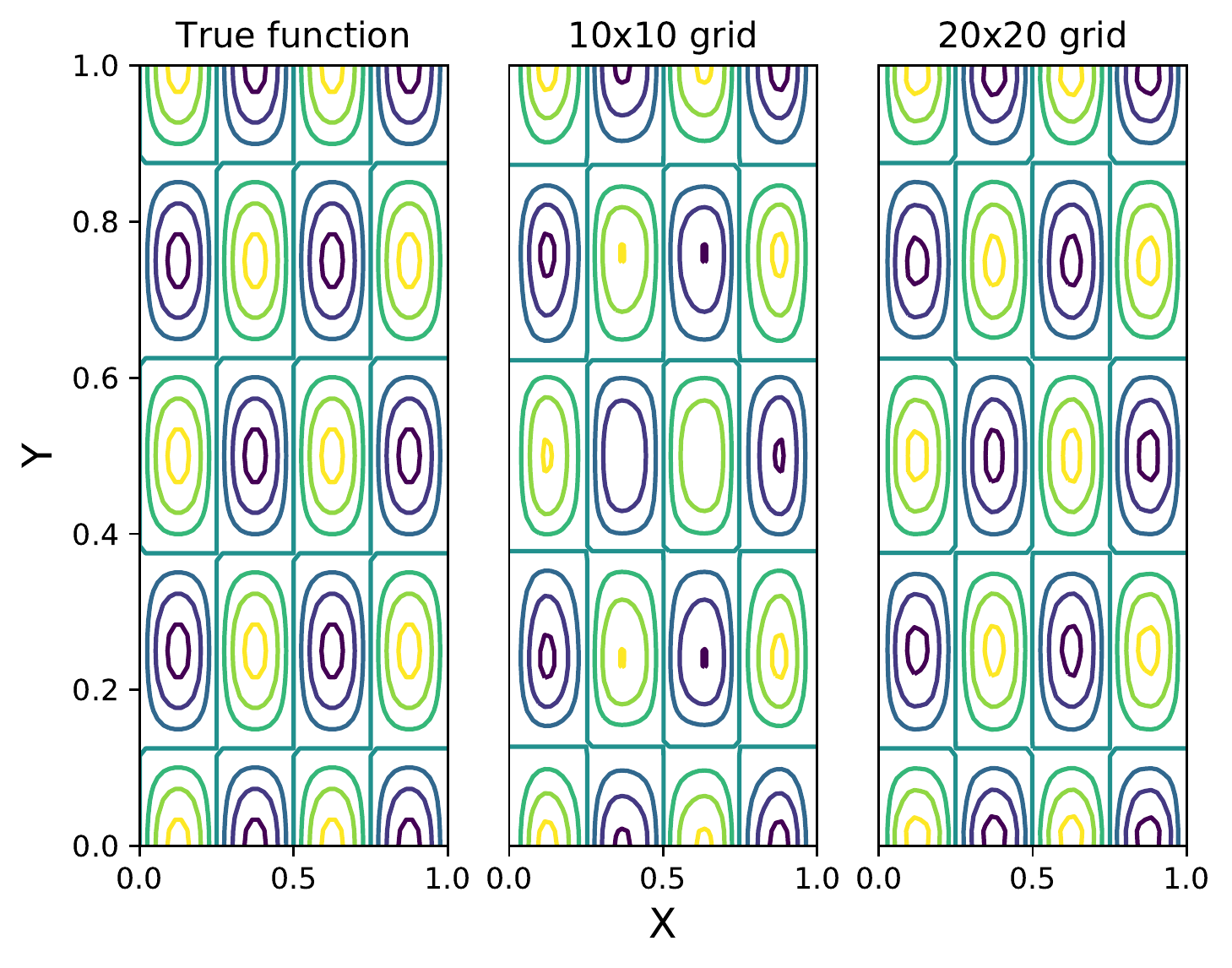}}
\caption{Two-dimensional $XY$ space grid used for regression with data generated by function $f(x,y) = \sin(4\pi x)\cos(4\pi y)$. The function values are represented by the color code. (a) The true function. (b) The regression result using 10$\times$10 grid. (c) 20$\times$20 grid.}
\label{Regression2D}
\end{figure}

For higher dimensional input ${\bf x}\in\mathbb{R}^d$, the squared exponential kernel function as an instance of a product kernel allows a compact and efficient representation of kernel matrix, namely ~\cite{saatcci2012scalable} 
\begin{equation}
    K = K^{(1)} \otimes K^{(2)} \otimes\cdots \otimes K^{(d)}
\end{equation} where the matrix element associated with the $n$th dimension is given by $K^{(n)}_{\bf x, \bf x'} = k(x_n,x'_n)$. Fig.~\ref{Regression2D} shows the regression results for data generated from the function $f(x,y)=\sin(4\pi x)\cos(4\pi y)$ using the 10x10 (middle) and 20x20 (right) grid, respectively, and $\ell = 0.03$. 

\section{Kernel Matrix Reconstruction}\label{sec:recons}

Thus far it has been seen how the eigen decomposition of kernel matrix using the standing wave basis in (\ref{EigenBasis}) leads to the GP prediction given one-dimensional (Figure~\ref{Regression}) and two-dimensional grid inputs (Figure~\ref{Regression2D}). Although the size of grid and the ensuing length $\Delta$ can be varied arbitrarily, the length scale always has a constraint. More precisely, demanding the positive semi-definiteness of kernel $K$, which is guaranteed if the minimum eigenvalue $\lambda$ in (\ref{eigenvalue}) is positive, leads to the constraint $\ell/\Delta < 1/\sqrt{2\ln 2}\approx0.85$. However, an exact GP in which every correlation term is retained is free of such constraint on $\ell$. In the following, we show how to incrementally relax the constraint by reconstructing the kernel matrix.

\begin{figure}[ht]
\centerline{\includegraphics[width=0.8\columnwidth]{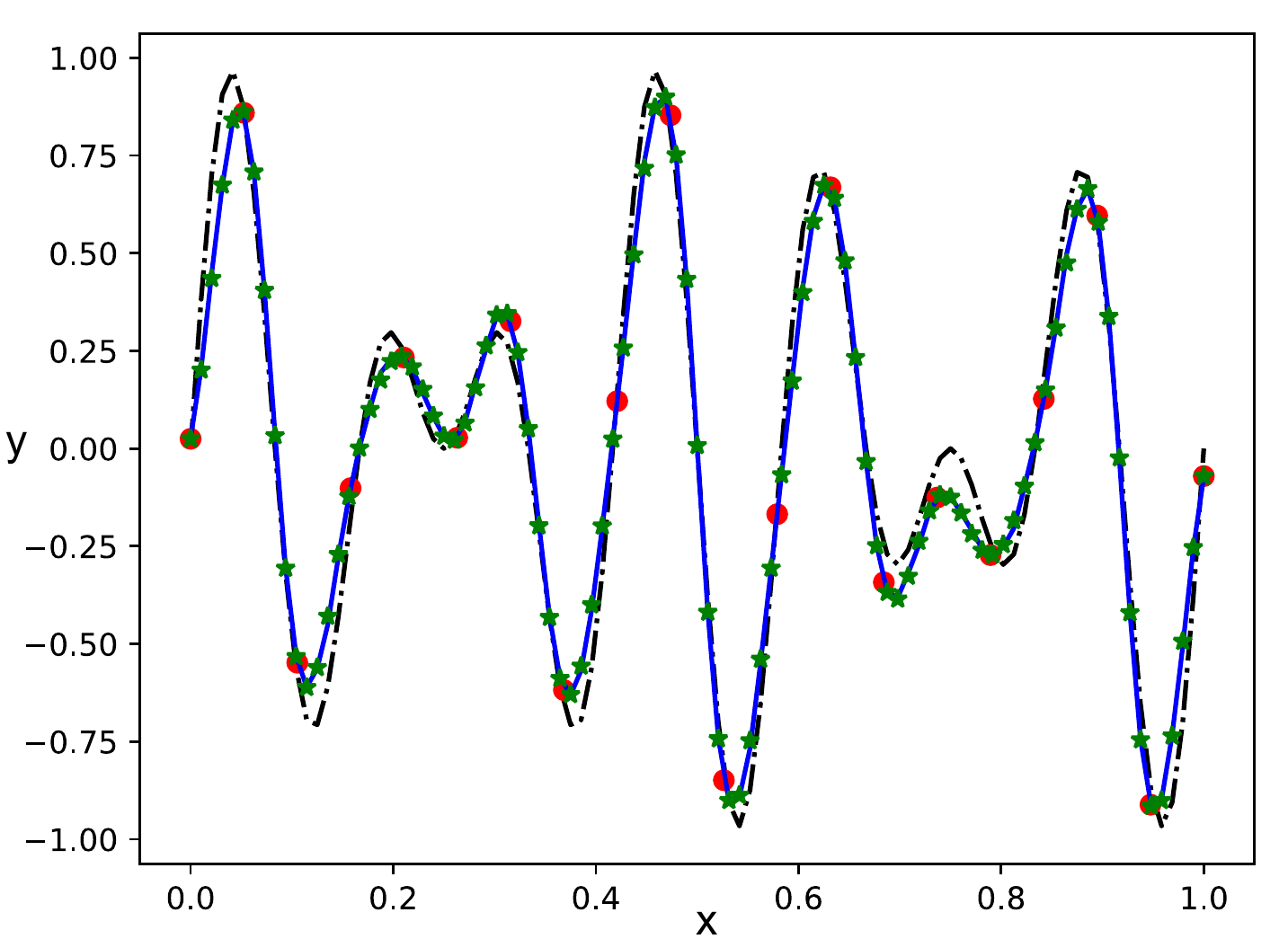}}
\centerline{\includegraphics[width=0.8\columnwidth]{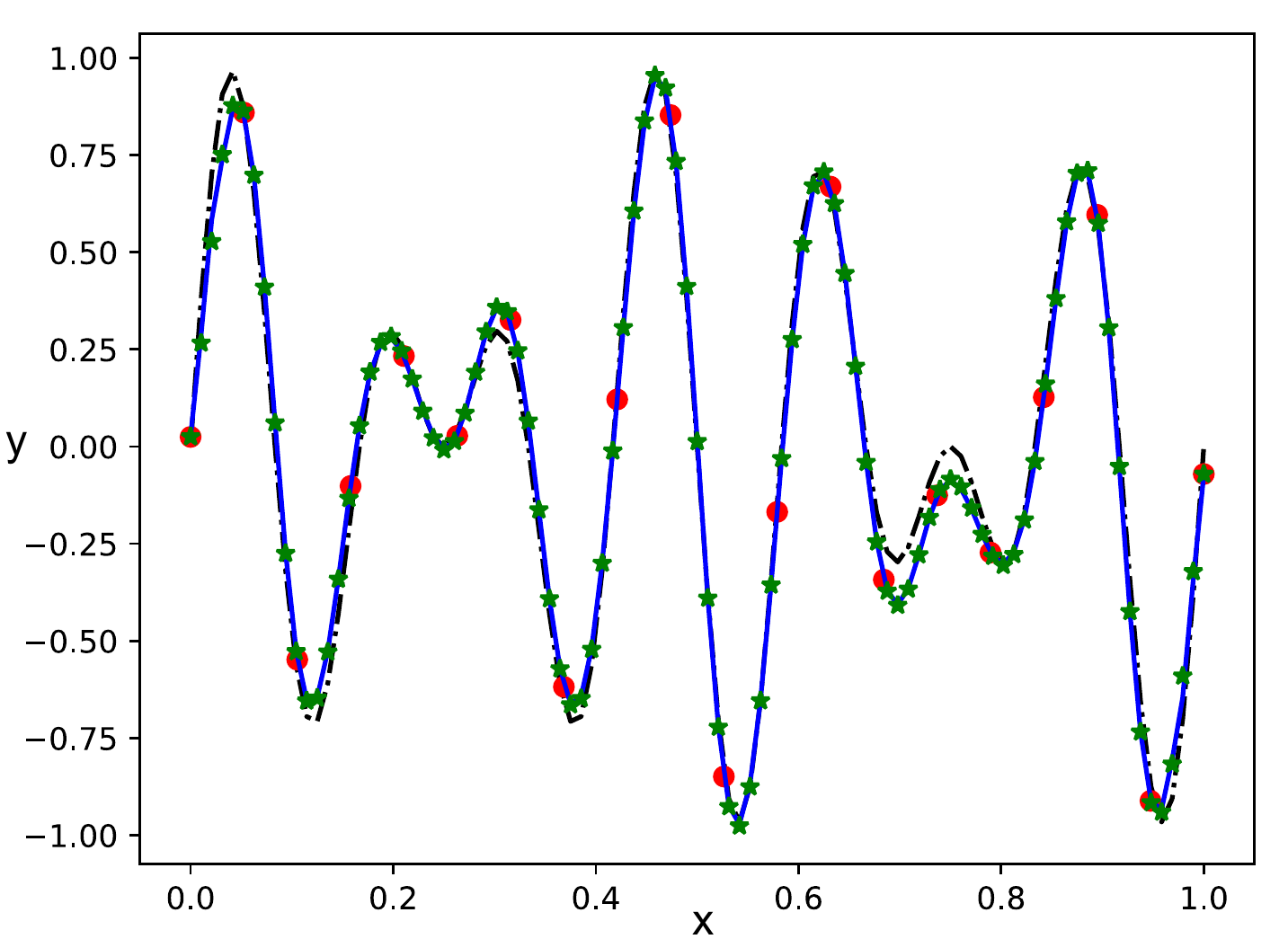}}
\caption{Regression results from the SWD-GP (blue lines) and exact GP (green star symbols) in comparison with the true function (black-dashed lines). Red dots denote the training data points generated by $f(x) = \cos(2\pi x)\sin(12\pi x)$ (black dashed line) on the grid of 20 points in [0, 1]. (a) Fit is obtained by applying tridiagonal SWD-GP and use length scale $\ell=$ 0.03. (b) Fit is obtained from the reconstructed pentadiagonal kernel matrix with $\ell=$ 0.04. 
}
\label{RegressionReconstructed}
\end{figure}

As a motivation, let us examine Fig.~\ref{RegressionReconstructed}(a) where the results from exact GP (green star) and tridiagonal SWD-GP (blue solid line) for  data generated by $y=\sin(12\pi x)\cos(2\pi x)$ (black dashed line) using $\ell/\Delta = 0.57$ for a grid of 20 points in [0,1] are shown. First, we note that the GP and the present SWD-GP have identical results, suggesting that the SWD-GP is indeed a good approximation. Secondly, both results here are unable to capture the curvature feature near the extreme points, e.g. at $x=1/8$ there is a subtle discrepancy against the true function. Besides the fact that there are not sufficiently many data points near these points, increasing the length scale may work, as suggested by the exact GP result (green star) in Fig.~\ref{RegressionReconstructed}(b) where a longer $\ell/\Delta=0.76$ is used. However, our simulation shows that directly applying the tridiagonal SWD-GP with $\ell/\Delta=0.76$ does not generate as a smooth prediction mean as the case with $\ell/\Delta=0.57$ does. This is because as the length scale approaches the upper bound, the smallest eigenvalue approaches zero and creates numerical instabilities. 


In order to extend the applicable range for $\ell/\Delta$, we are interested in restoring the next nearest neighbor (NNN) correlation terms proportional to $\alpha^4$ in $K$, i.e. the matrix elements $K_{ij}$ with $|i-j|=2$. Inspection of (\ref{eig_eq}) suggests that the eigenvalue acquires a new contribution so that the new eigenvalue $\lambda^{(1)}$ reads
\begin{equation}\label{updated_eig_value}
    \lambda^{(1)}/\sigma^2 = 1 + 2\alpha\cos\theta + 2\alpha^4\cos2\theta 
\end{equation} while the eigenvectors are left intact. In fact, neither the eigenvalues nor the eigenvectors are exact because the correlation among the grid points is not symmetric near the boundary.

Instead of solving for the true eigenvalues and true eigenvectors of the kernel matrix $K$ containing the NNN correlation, we reconstruct the kernel matrix which is still diagonalized by the eigen basis in (\ref{EigenBasis}) and has eigenvalues as in (\ref{updated_eig_value}). 
We apply the following identity to reconstruct $\tilde K$,
\begin{equation}\label{series}
\begin{split}
    \sum_{k=1}^M\cos(q\theta_k)[{\rm v}_{k}{\rm v}_{k}^T]_{ij} = \frac{1}{2(M+1)}[a(i-j+q)+\\
    a(i-j-q)-a(i+j+q)-a(i+j-q)]
\end{split}
\end{equation} with the coefficient defined as,
\begin{equation}
    a(p)\equiv\sum_{k=1}^M\cos \frac{kp\ \pi}{M+1} = \frac{-1-(-1)^p}{2}\:,
\end{equation} for $p\in{\mathbb Z}_{2N+2}\backslash \{0\}$ and $M$ otherwise. It is easy to see that the  constant ($q=0$) together with the first cosine term ($q=1$) in (\ref{updated_eig_value}) gives rise to the original tridiagonal matrix. As for the last term in (\ref{updated_eig_value}) corresponding to $q=2$, the RHS of (\ref{series}) associated with $i-j=\pm 2$ equal $\frac{1}{2}$, which leads to the desired NNN correlation term. However, special care is needed for the cases for $i+j-q=0$ and $i+j+q = 2N+2$, which, respectively, correspond to $(i,j)=(1,1)$ and $(i,j)=(M,M)$ with $q=2$. It is easily verified that the reconstructed kernel matrix elements corresponding to the end points in grid are $\tilde K_{11} = \tilde K_{MM} = K_{11}-\sigma^2\alpha^4$.  


In summary, the reconstructed pentadiagonal kernel matrix $\tilde K$, which contains the NNN correlation terms, reads,
\begin{equation}\label{eq:kij_mod}
    \tilde K_{ij} = K_{ij} + \sigma^2\alpha^4\left[\delta_{|i-j|,2}-\delta_{i,j}(\delta_{i,1}+\delta_{i,M})\right]\:.
\end{equation} As shown in Fig.~\ref{RegressionReconstructed}(b), we employ the pentadiagonal $\tilde K$ using $\ell/\Delta = 0.76$ to repeat the regression task, and we are able to reproduce the same result as the exact GP with same length scale. Moreover, including NNN correlation also improves the accuracy near the local extreme points.


\section{Latent Grid SWD-GP}\label{sec:latent}

\subsection{Bayesian Data Projection} 

When applying to data $({\bf X, y})$ where the input points $\bf X$ are randomly selected, we shall construct a latent grid $\bf X_g$ so that the previous SWD-GP approach can be employed. The missing components are the corresponding function values $\bf g$ evaluated at these grid points. Now we assume that $\bf g$ is a set of random variable and we find the most probable values $\bf{ \bar{g}}$ in the Baysian approach. We first calculate the posterior distribution associated with $\bf g$,
\begin{equation}
    p({\bf g}|{\bf X_g, X, y}) \propto \left(
    \prod_{i=1}^N p(y_i|{\bf x}_i,{\bf X_g, g})
    \right)p(\bf g|X_g)
\end{equation} where we have assumed independence among the given data. The likelihood function follows the usual GP predictive distribution by assuming $\bf g$ is known,
\begin{equation}
    p(y|{\bf x, X, g}) = \mathcal N(K_{\bf xg}K^{-1}_{\bf gg}{\bf g}, \sigma_{\bf x}^2)\:,
\end{equation} and the variance at $\bf x$ is denoted by
\begin{equation}\label{eq:sigma_x}
    \sigma_{\bf x}^2 = K_{\bf xx} - K_{\bf xg}K_{\bf gg}^{-1}K_{\bf gx} + \sigma_N^2\: 
\end{equation} 
with $\sigma_N^2$ being the variance of the observation noise. The prior has its zero mean distribution with the kernel matrix $K_{\bf gg}$, 
\begin{equation}
    p({\bf g|X_g}) = {\mathcal N}(0, K_{\bf gg})\:.
\end{equation} With some tedious calculations, the posterior distribution can be shown to be,
\begin{equation}
    p({\bf g}|{\bf X_g, X, y}) = \mathcal N(\bar{\bf g}, A^{-1})\:,
\end{equation} where the updated variance is encoded in 
\begin{equation}
    A^{-1} = K_{\bf gg}Q^{-1}K_{\bf gg}\:,
\end{equation} with the matrix
\begin{equation}\label{matrix_Q}
    Q = K_{\bf gg}+K_{\bf gx}\Lambda^{-1}K_{\bf xg}\:.   
\end{equation} The diagonal matrix $\Lambda = {\rm diag}(\sigma^2_{{\bf x}_1}, \sigma^2_{{\bf x}_2},\cdots,\sigma^2_{{\bf x}_N})$. Furthermore, the mean vector $\bf{\bar g}$ represents the most probable projected function values,
\begin{equation}\label{mostprobable_g}
    \bar{\bf g}=K_{\bf gg}Q^{-1}K_{\bf gx}\Lambda^{-1}\bf y\:.
\end{equation}


Now, by marginalizing the latent function values $\bf g$, we may obtain the predictive distribution for the function value $y_*$ associated with the test point $\bf x_*$,  
\begin{equation}\label{general_data_prediction}
    p(y_*|{\bf x}_*) = \mathcal N(K_{*\bf g}K^{-1}_{\bf gg}\bar{\bf g},\sigma_*^2+K_{*\bf g}Q^{-1}K_{\bf g*})\:,
\end{equation} which is consistent with the results presented in \cite{snelson2006sparse, quinonero2005unifying}.

\subsection{Perturbation and Matrix Inversion}

Unlike the grid data to which the inverted kernel matrix $K^{-1}_{\bf gg}$ is readily applicable, here we have to invert the matrix $Q$ in (\ref{matrix_Q}) in order to find the mean of predictive distribution in (\ref{general_data_prediction}). To approach this problem, we shall eventually write $Q$ in terms of the SW eigen basis of $K_{\bf gg}$. Hence, the second term in (\ref{matrix_Q}) is treated as perturbation so that the eigenvalues in (\ref{eigenvalue}) as well as eigenvectors in (\ref{EigenBasis}) are modified. Here, to make the calculation more efficient and accessible, we regroup the $M$ eigenvectors $\{{\rm v}_i\}$ with $i = 1, 2,\cdots,M$ into pairs $\{{\rm v}_i, {\rm v}_{M+1-i}\}$, and for ease of notation we use $\rm v$ and $\bar{{\rm v}}$ to represent the two eigenvectors in a given pair.
Projecting $Q$ onto the subspace spanned by $\rm v$ and $\bar{{\rm v}}$, we arrive at the following eigenvalue equation for each pair, and the associated eigenvalue $\chi$ and eigenvector $\rm u$ can be obtained by solving,
\begin{equation}\label{perturbation_eq}
\left(\begin{array}{cc}
    \lambda + \epsilon & \delta \\
    \delta & \bar\lambda + \bar\epsilon
\end{array}\right)\rm u = \chi \rm u\:.   
\end{equation} The value $\epsilon$ is the first-order correction to unperturbed eigenvalue $\lambda$ due to the term ${\rm v}^TQ{\rm v}$, while $\delta$ is the cross term from ${\bar{\rm v}}^TQ{\rm v}$. It can be shown that the updated eigenvalues and eigenvectors are given by,
\begin{equation}
    \chi_{\pm} = \beta_+\pm\sqrt{\beta_-^2+\delta^2}\:,
\end{equation} and
\begin{equation}
\left(\begin{array}{cc}
    \rm u_+ \\
    \rm u_-
\end{array}\right) = 
\left(\begin{array}{cc}
    \cos\phi & \sin\phi \\
    -\sin\phi & \cos\phi
\end{array}\right)
\left(\begin{array}{cc}
    \rm v \\
    {\bar{\rm v}}
\end{array}\right)
\end{equation} with the rotation angle $\phi$ specified by
\begin{equation}
    \tan\phi=\frac{\sqrt{\beta_-^2+\delta^2}+\beta_-}{\sqrt{\beta_-^2+\delta^2}-\beta_-}\:.   
\end{equation} The parameter $\beta_{\pm} = [(\lambda+\epsilon)\pm(\bar\lambda+\bar\epsilon)]/2$. Consequently, the matrix $Q$ has the following decomposition,
\begin{equation}\label{eq:eigencomp}
    Q^{-1} = \sum_{i=1}^M \frac{{\rm u}_i{\rm u}_i^T}{\chi_i}\:.
\end{equation}

\begin{figure}[h]
\centerline{\includegraphics[width=0.5\columnwidth]{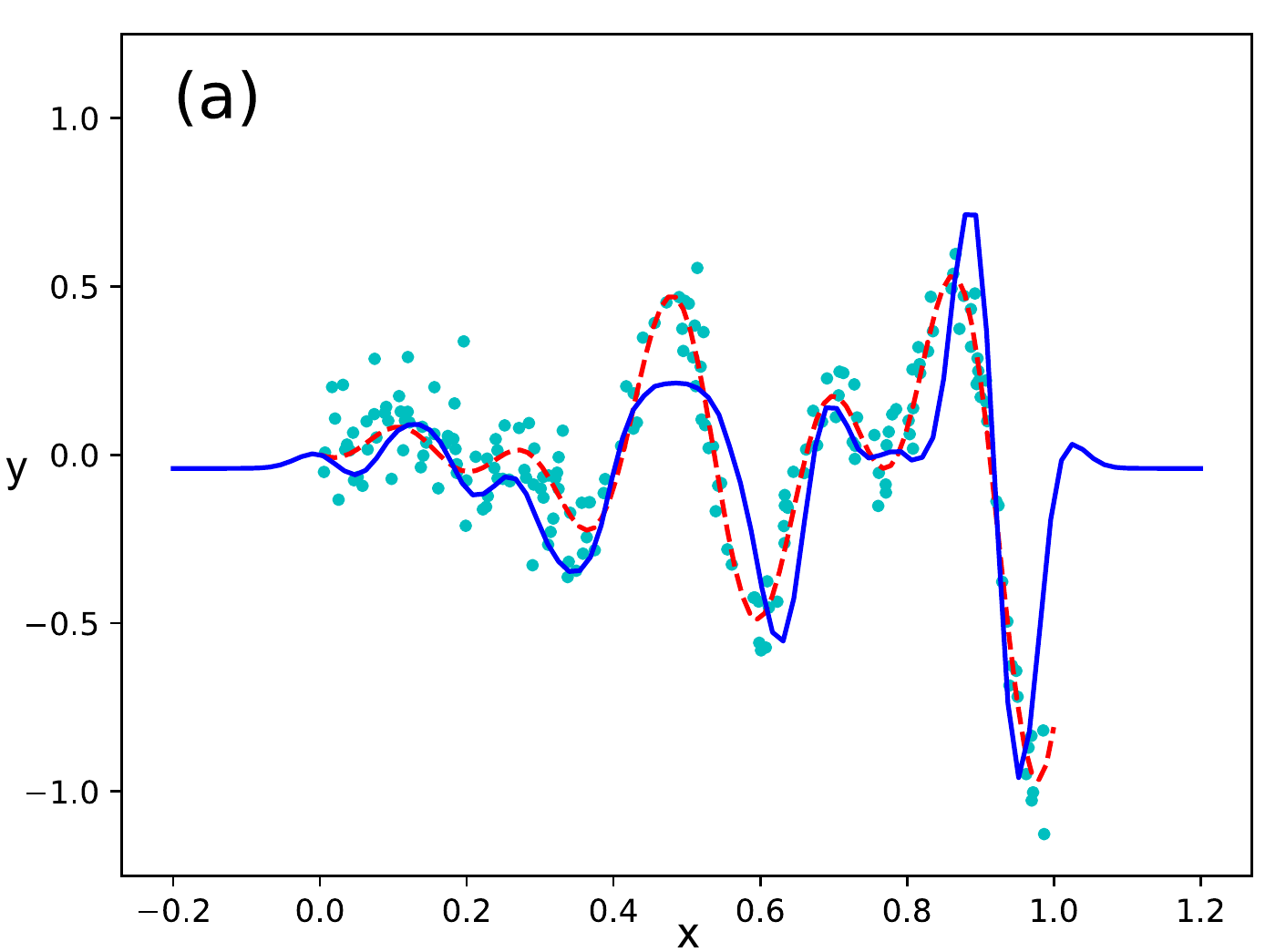}
\includegraphics[width=0.5\columnwidth]{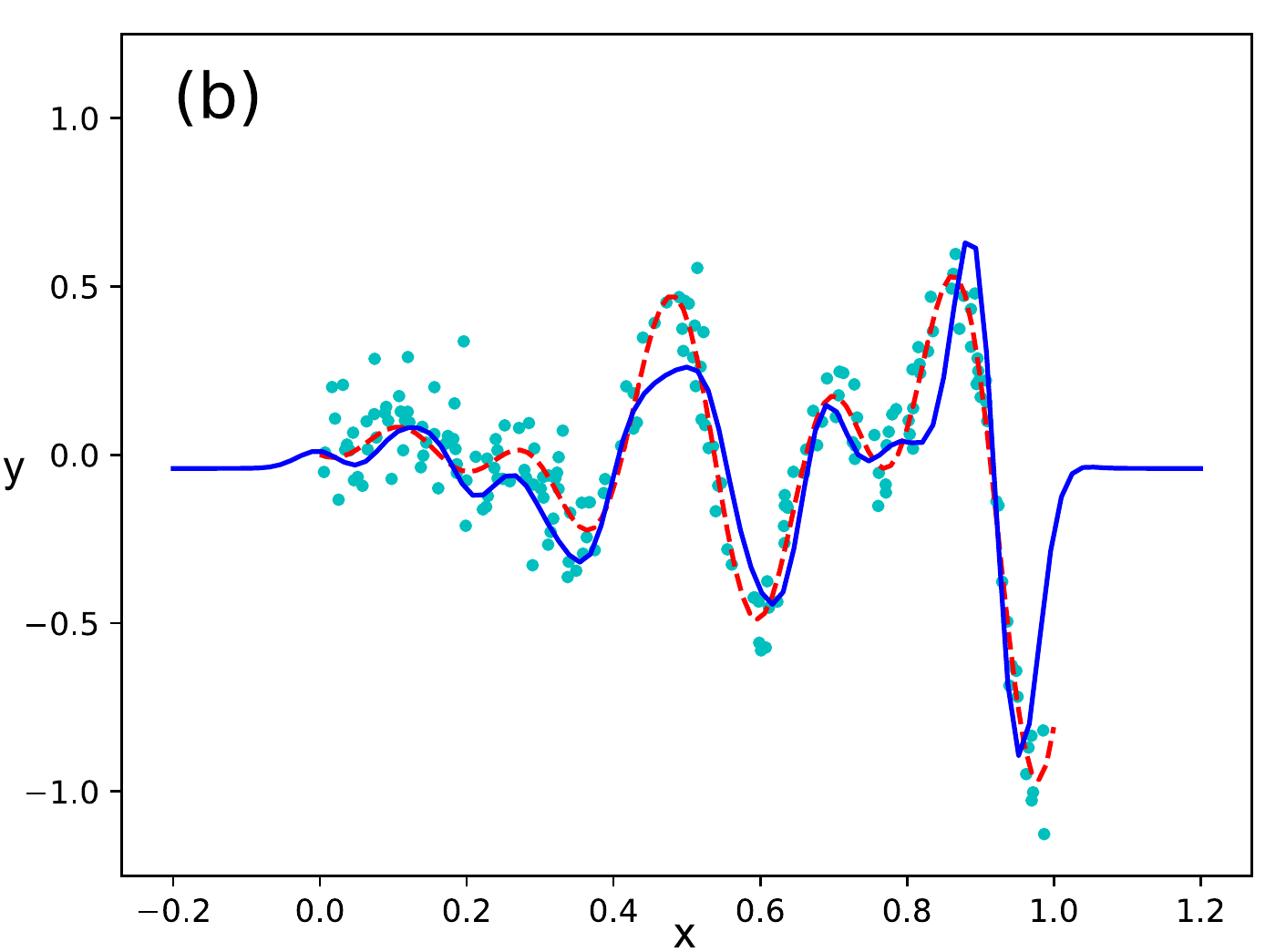}}
\centerline{\includegraphics[width=0.5\columnwidth]{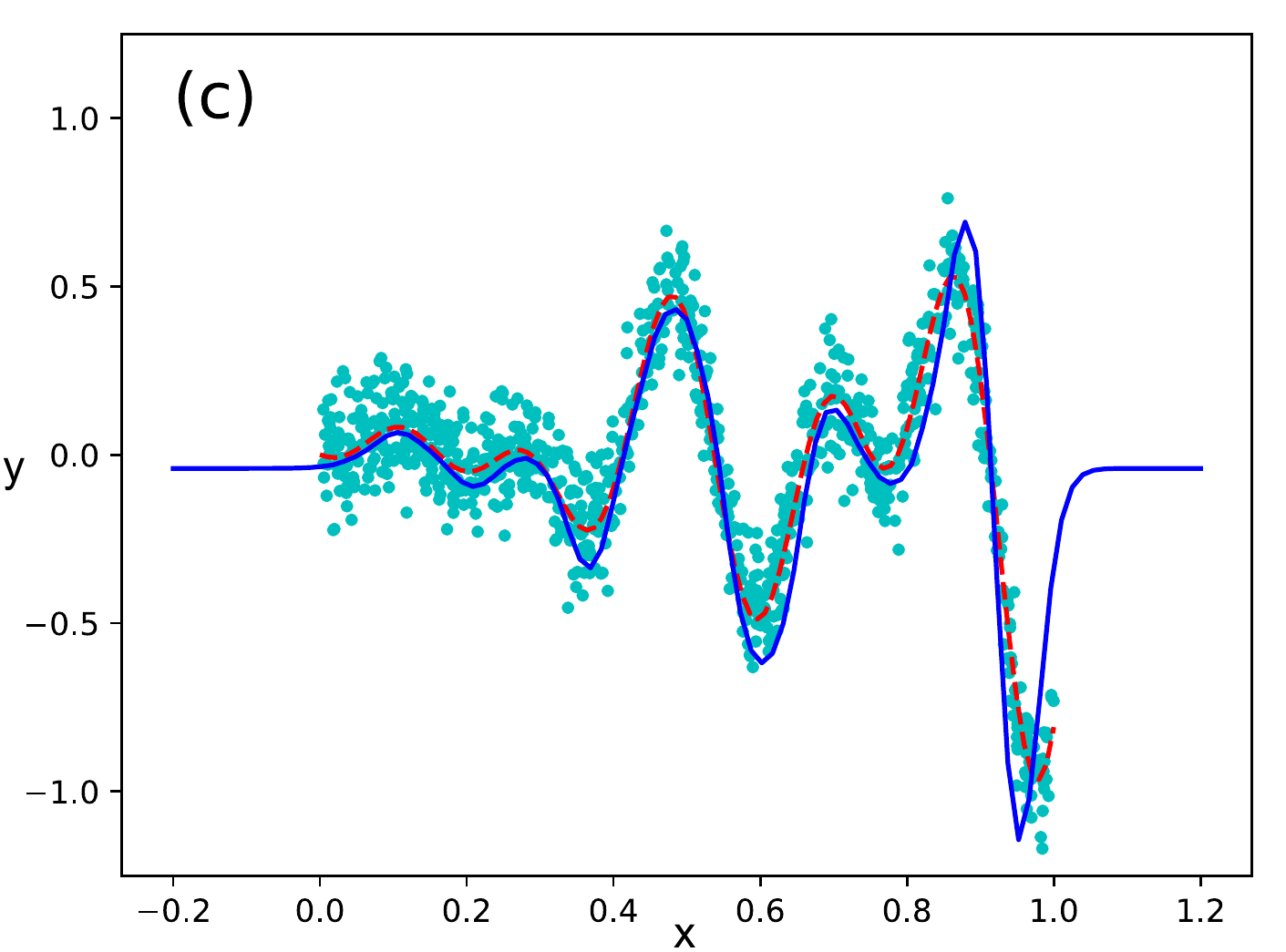}
\includegraphics[width=0.5\columnwidth]{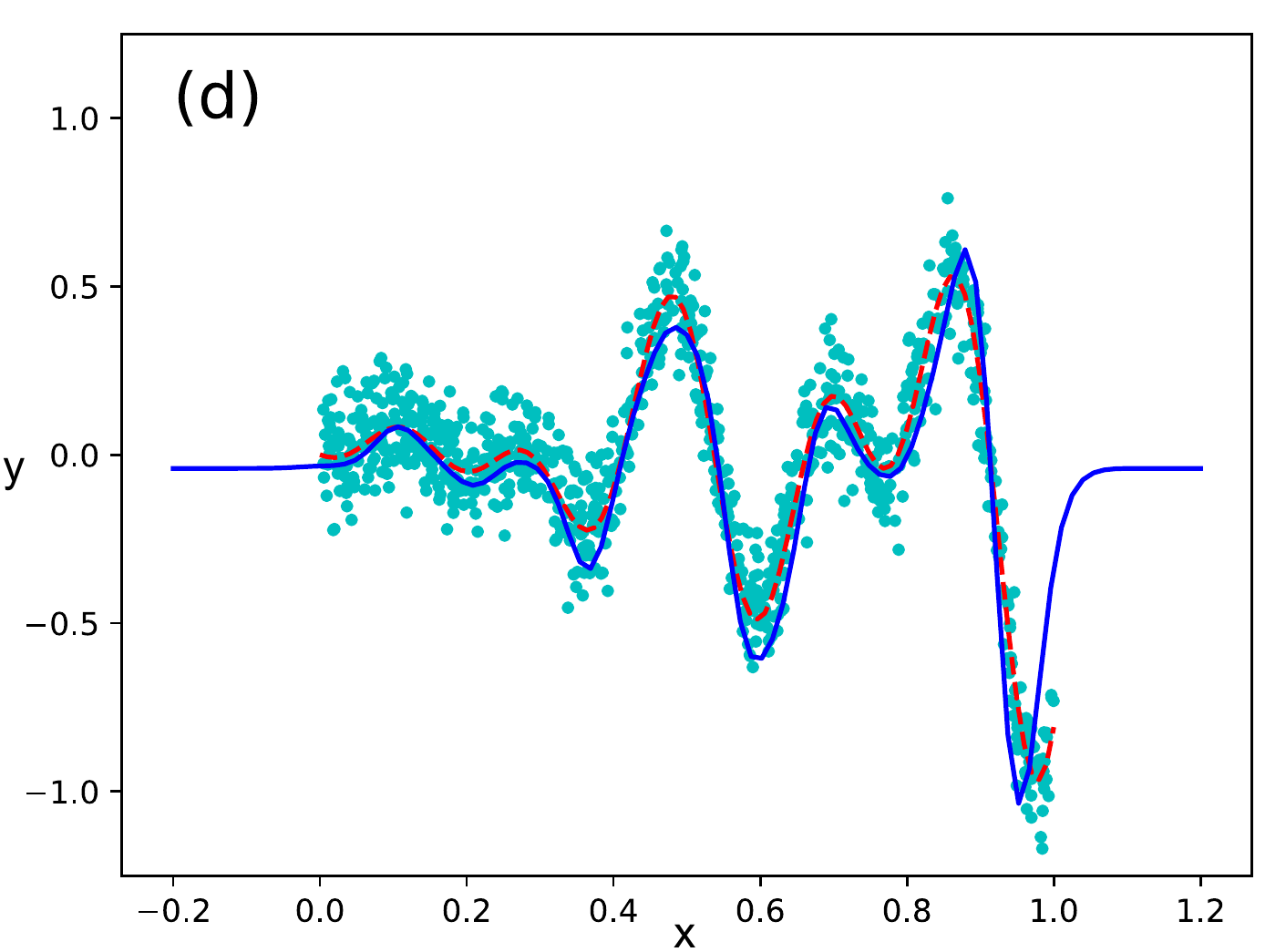}}
\caption{Illustration of latent grid SWD-GP 1st order $\delta=0$ in (\ref{perturbation_eq})] and 2nd order ($\delta>0$) perturbation using data generated from function $f(x)=x\cos[8\pi (x+0.15)]\cos(2\pi x)$ (red dashed line) plus noise. A latent grid of 20 points and $\ell = 0.03$ are used, and the prediction mean is denoted by the blue solid line. (a) First-order purturbation with 200 data points. (b) Second-order perturbation with same parameters as (a). (c) Same as (a) but with 1000 data points. (d) Same as (b) but the number of data points increased to 1000. 
} 
\label{GeneralRegression}
\end{figure}

Next, we apply the the above approach to the data generated by the function $y=x\cos[8\pi(x+0.15)]\cos(2\pi x)$ 
with input $x$ randomly selected from [0,1]. Figure~\ref{GeneralRegression}(a)--(b) and (c)--(d) show the results for the training number $|X|=$ 200 and 1000, respectively. Panels (a) and (c) in Fig.~\ref{GeneralRegression} are obtained in first-order perturbation, or by setting $\delta = 0$ in (\ref{perturbation_eq}). With a larger number of training points [Fig.~\ref{GeneralRegression}(c) versus (a)], the most probable projected function values $\bar{\bf g}$ are closer to the true function values on the grid. Panels (b) and (d) correspond to a fuller second-order perturbation treatment by considering the eigenvector correction due to nonzero $\delta$ in (\ref{perturbation_eq}). Incremental improvements, and in particular near the end points, can be seen by comparing (a) versus (b), or (c) versus (d).

\begin{figure}[ht]
\centerline{\includegraphics[width=\columnwidth]{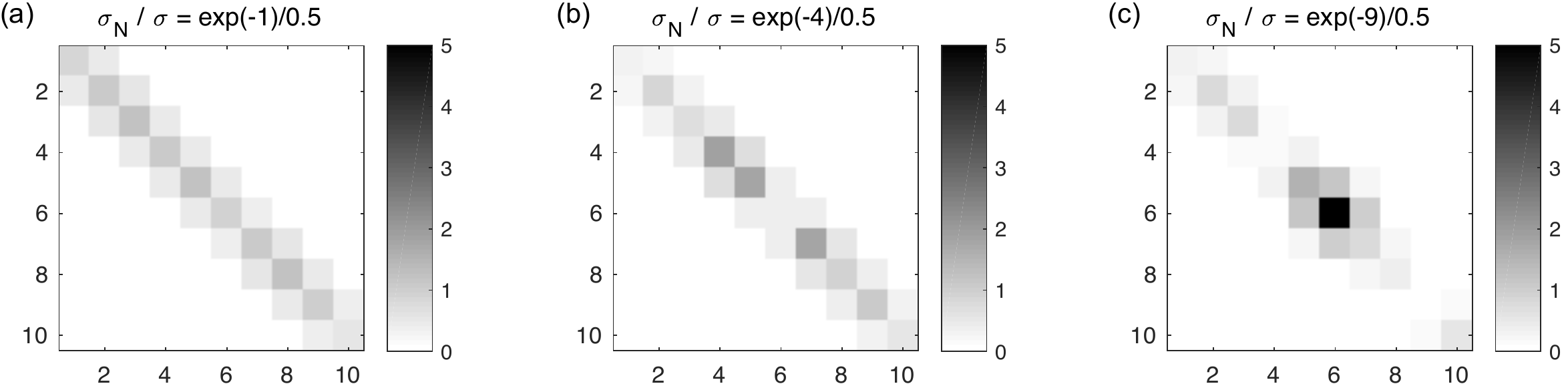}}
\centerline{\includegraphics[width=\columnwidth]{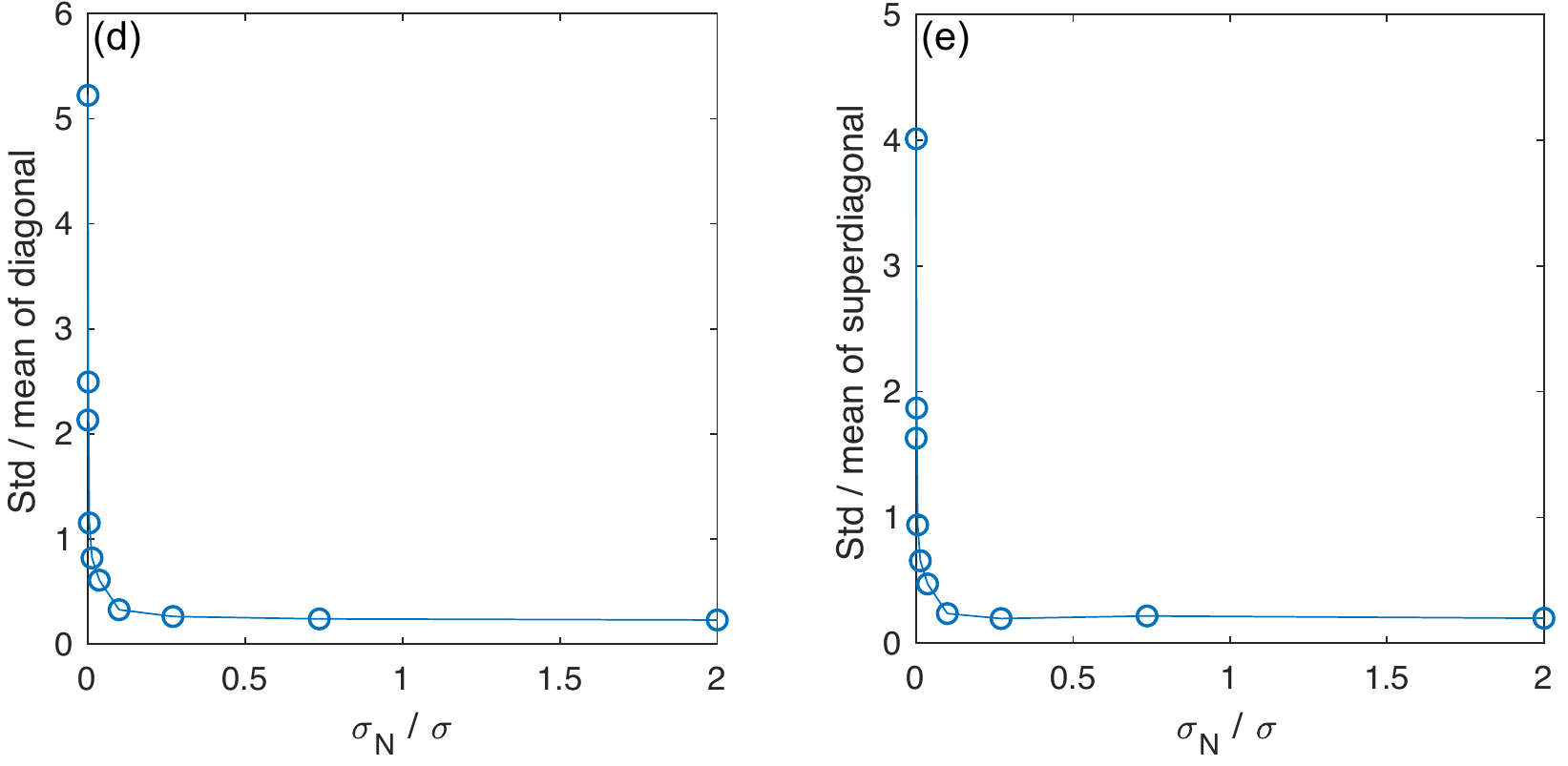}}
\caption{Variation of matrix elements of $K_{\rm {gx}}\Lambda^{-1}K_{\rm {xg}}$, the second matrix in Eq.~(\ref{matrix_Q}). Upper panel: results for grid size $|{\bf X_g}| = 10$ and number of training points $|{\bf X}| = 200$ with different noise parameters $\sigma_N/\sigma$. From (a) to (c), decreasing $\sigma_N/\sigma$'s leads to stronger variations among the matrix elements along the diagonal, superdiagonal, and subdiagonal. Lower panel: Using $|{\bf X_g}| = 50$ and $|{\bf X}| = 1000$, the ratio of standard deviation over the mean for matrix elements along the diagonal (d) and sub/super diagonal (e) against $\sigma_N/\sigma$ is obtained. A critical value $\sigma_N/\sigma = 0.1$ is suggested for valid perturbation. 
} 
\label{matrixvariation}
\end{figure}

The above results suggest that the inference through Eq.~(\ref{mostprobable_g}) using the approximated matrix inversion is effective for larger $|X|$ and noise parameter $\sigma_N/\sigma$ of order $10^{-1}$. To get insight into why it works, we shall show that the second matrix in Eq.~(\ref{matrix_Q}) can be approximated as a linear combination of identity and super/sub diagonal matrices,  
\be\label{residual}
    [K_{\rm {gx}}\Lambda^{-1}K_{\rm {xg}}]_{ij} = a\delta_{ij}+b\delta_{|i-j|=1} + R_{ij}\:,
\ee where $a$ and $b$ stand for the averages of diagonal and super/sub diagonal matrix elements, respectively. The last term $R$ denotes the residual matrix. Therefore, the first two terms in Eq.~(\ref{residual}) commute with $K_{\bf gg}$ while $R$ is $\it small$ if the variations among the diagonal and super/sub diagonal matrix elements are weak. We demonstrate in Fig.~\ref{matrixvariation}(a)--(c) that larger values of $\sigma_N/\sigma$ result in weaker variation among the matrix elements. This is further confirmed in panels (d) and (e) where the ratio of standard deviation over mean along diagonal [Fig.~\ref{matrixvariation}(d)] and super/sub diagonal [Fig.~\ref{matrixvariation}(e)] lines are sufficiently small for $\sigma_N/\sigma > 0.1$. 

\section{Simulations}\label{sec:simu}

We compare the method presented here with three existing, popular methods---exact GP, fully independent training conditional GP (FITC-GP) \cite{snelson2006sparse,quinonero2005unifying}, and Kernel Interpolation for Scalable Structured GP (KISS-GP) \cite{wilson2015kernel}. The main quantities of interest are data efficiency---the amount of data needed to reach a certain accuracy---and run time.
We apply all methods to a toy regression problem, where 
the x-values are drawn uniformly random from $[0,1]$, and the corresponding y-values are given by $y=\sin(\frac{5\pi}{x+0.1}) + {\rm Normal}(0, 0.2^2)$.
Accuracy is measured by standardized mean-square error (SMSE). 
Run time is measured on a 2015 model of MacBook with 1.1 GHz dualcore Intel Core M and 8GB 1600MHz LPDDR3.
All methods are implemented in MATLAB 2017b. We largely followed an example\footnote{www.gaussianprocess.org/gpml/code/matlab/doc/demoGrid1d.m} in the documentation of the GPML toolbox for using the existing methods. For the implementation of our latent grid standing wave decomposition GP (LGSWD-GP), we used only first-order correction on the eigenvalues (i.e., setting $\delta=0$ in \eqref{perturbation_eq}), since Fig.~\ref{GeneralRegression} suggests that the improvement from adding second-order correction on the eigenvectors is relatively small compared to increasing the number of training points. Our code is available at TBA.\footnote{ https://github.com/CoDaS-Lab/LG-SWD-GP}

\begin{figure}[ht]
\includegraphics[width=\columnwidth]{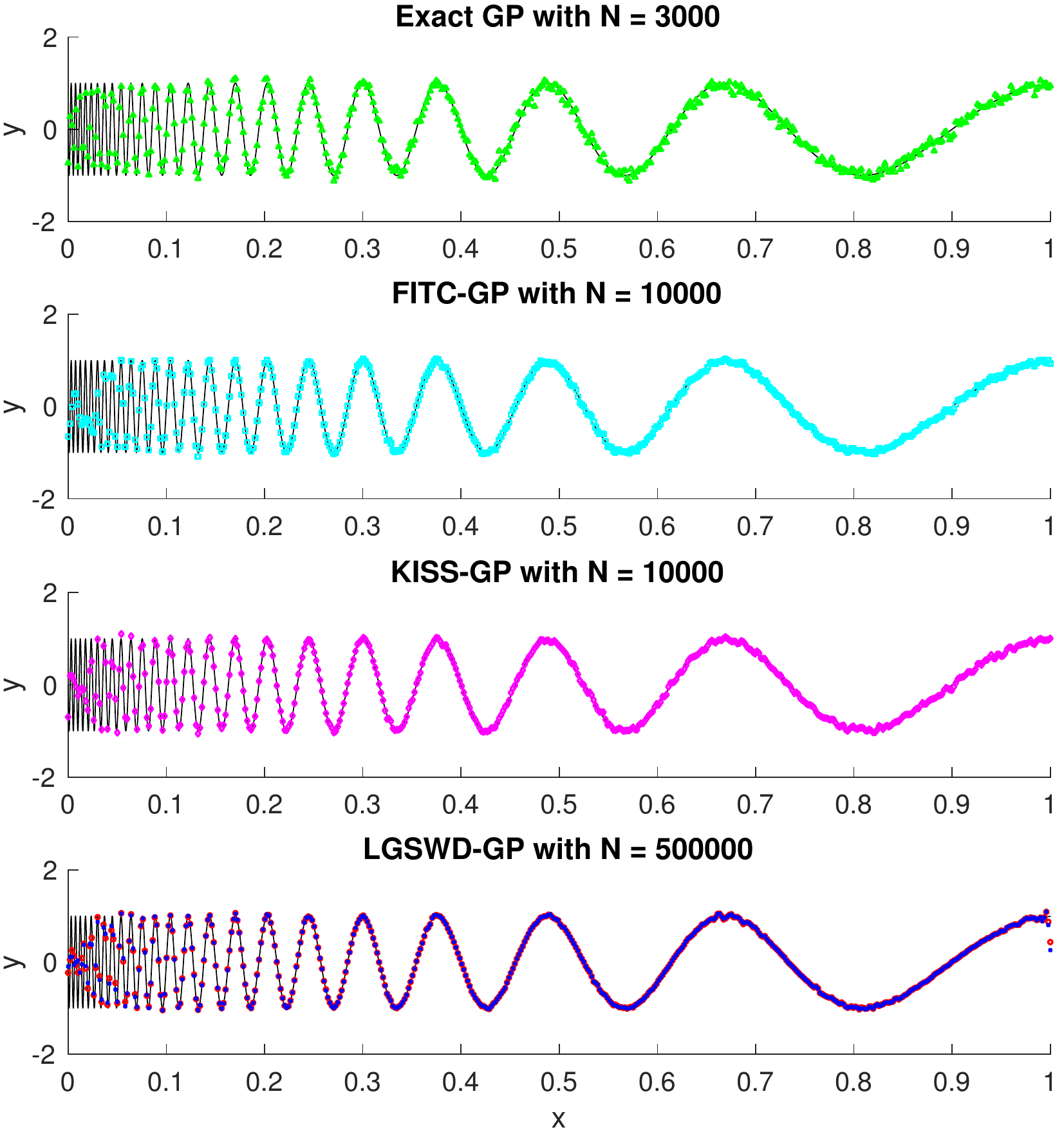}
\caption{Example fits of (a) exact GP with $N=3000$, (b) FITC-GP with $N=10^4$, (c) KISS-GP with $N=10^4$, and (d) our LGSWD-GP with $N=5\times10^5$, where $N$ is the number of training data points. The different $N$ used in each method corresponds to the largest $N$ used for that method in Fig.~\ref{fig:simu_err_time}. Black line represents the true underlying function; markers represent the mean of the predictive distribution at the test points. For (d), red circles and blue crosses correspond to tridiagonal and pentadiagonal reconstruction, respectively.
}
\label{fig:simu_fit}
\end{figure}

For all methods, the squared-exponential kernel is used. The kernel's overall variance, $\sigma^2$, is fixed at $0.5^2$, and the variance of likelihood is set to the true value, $0.2^2$. Exact GP and KISS-GP are allowed to optimize for the kernel's length scale, and FITC-GP is allowed to optimize for the location of the inducing points in addition to the kernel's length scale. These hyper-parameter optimization are limited to 20 iterations. Note that for our LGSWD-GP, the length scale can be determined without invoking hyper-parameter optimization (see Sec.~\ref{sec:recons}). The number of inducing points are fixed at $300$ throughout for FITC-GP, KISS-GP, and LGSWD-GP. The inducing grid in KISS-GP and LGSWD-GP are evenly spaced in $[0,1]$. The test points are $500$ evenly spaced points in $[0,1]$. 
Example fits are shown in Figure~\ref{fig:simu_fit}. 

\begin{figure}[ht]
\includegraphics[width=\columnwidth]{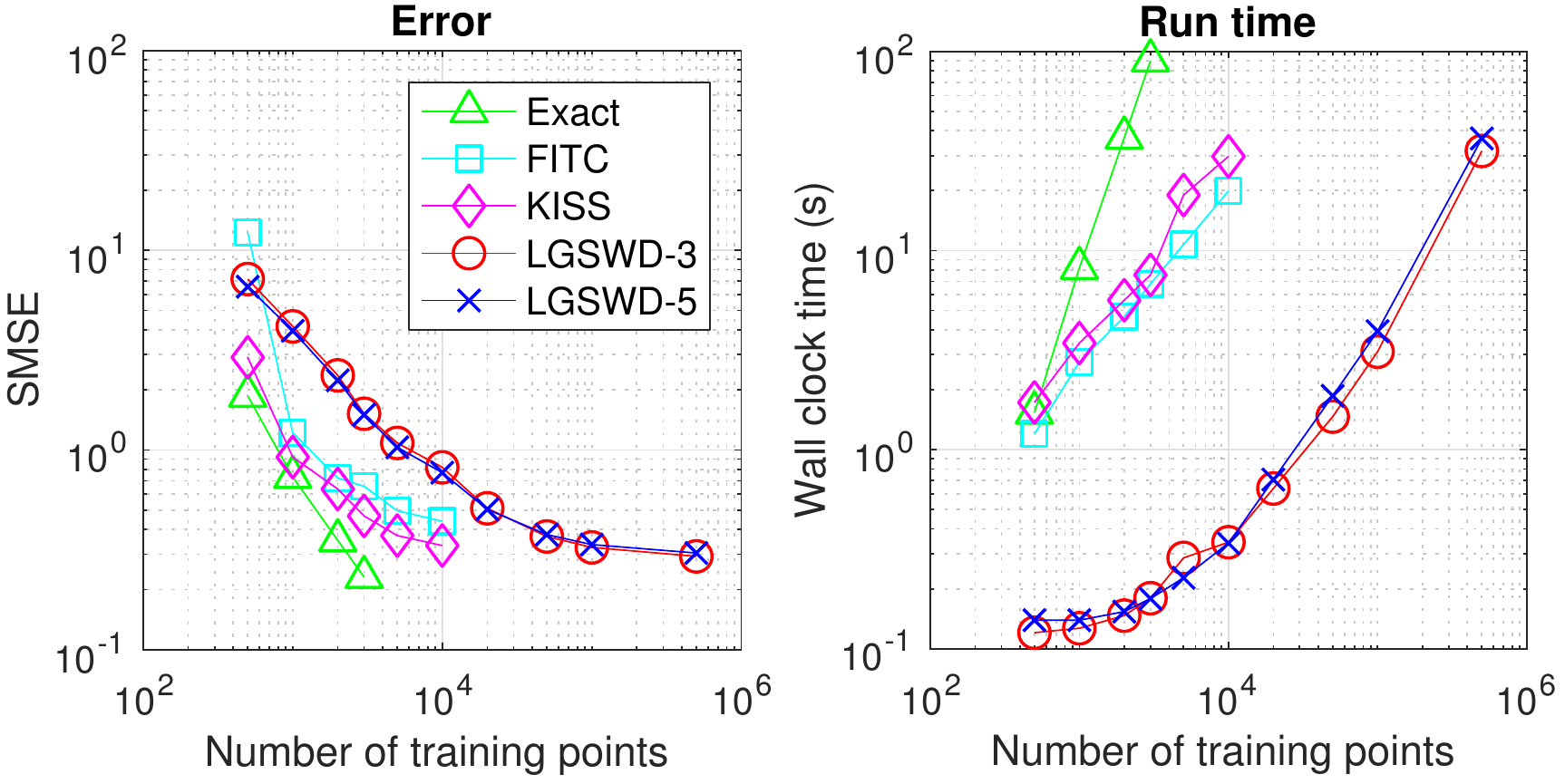}
\caption{SMSE error (a) and run time (b) as a function of number of training points for exact GP, FITC-GP, KISS-GP, and LGSWD-GP. LGSWD-3 and LGSWD-5 correspond to the tridiagonal and pentadiagonal approximations, respectively.}
\label{fig:simu_err_time}
\end{figure}

Figure~\ref{fig:simu_err_time} shows the accuracy and run time as a function of number of training points for the different methods. From the error plot, we see that our method has a lower data efficiency compared to existing methods; that is, compared to those methods, the LGSWD-GP needs more data to achieve the same accuracy.
Nevertheless, the run-time plot shows that the poorer data-efficiency is offset by the radical increase in computational efficiency.
The LGSWD-GP is roughly two orders of magnitude faster than FITC-GP and KISS-GP and is capable of reaching or even surpassing these methods in accuracy in the same amount of time.
A trade-off between speed and accuracy is expected. What is interesting is that these seemingly bold tridiagonal and pentadiagonal kernel approximations provides enough data efficiency to make the trade off favorable.

\begin{figure}[ht]
\includegraphics[width=\columnwidth]{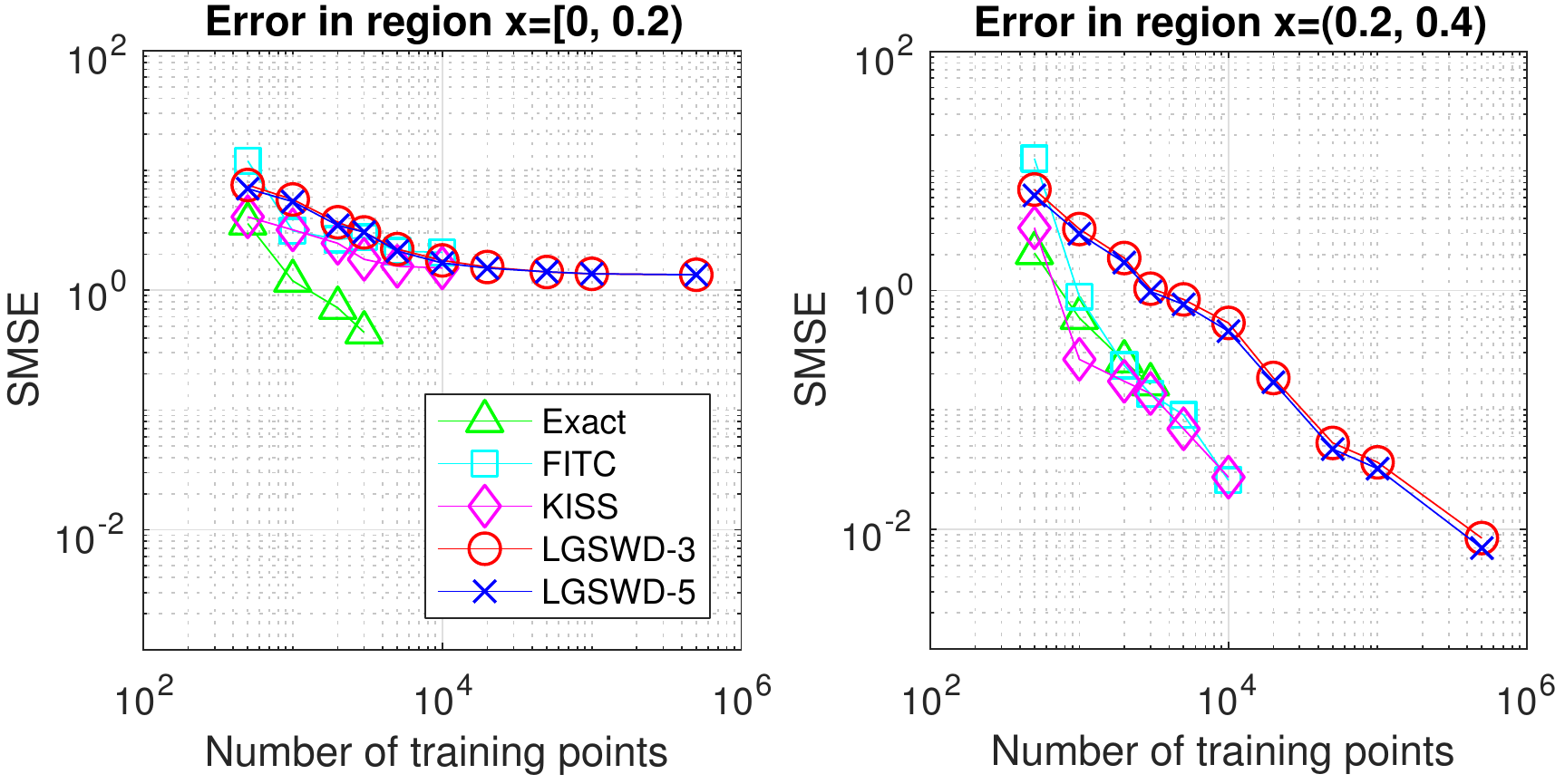}
\caption{Accuracy by regions. SMSE is calculated for the $100$ test points in regions (a) $x=[0,0.2)$ and (b) $x=(0.2,0.4)$. In theory, all the training points affect the prediction at every test point; thus, the x-axis is the same as that in Fig.~\ref{fig:simu_err_time}. But effectively, only roughly one fifth of the training points (those in the test region) contributed to the fit. The error plots for the remaining regions $(0.4, 0.6)$, $(0.6, 0.8)$, and $(0.8, 1]$ all look similar to (b).}
\label{fig:simu_err_time_p}
\end{figure}

Next, we analyze the accuracy by breaking the test region into five equi-sized regions, i.e., $[0, 0.2),(0.2, 0.4),\ldots,(0.8,1]$. 
Note that the effective length scale of the underlying function is different in each region (see Fig.~\ref{fig:simu_fit}). 
Let the effective length scale in region $r$ be $\xi_r$ and the number of training points in $\xi_r$ be $n_r = N\frac{\xi_r}{L}$.
The ratio $\Delta/\xi_r$  and $n_r$ are useful in characterizing the accuracy behavior of GP. 
First, all methods do relatively poorly when $n_r$ is small (Fig.~\ref{fig:simu_err_time_p}(a)) and relatively well when $n_r$ is large enough (Fig.~\ref{fig:simu_err_time_p}(b)), which is just a restatement of classical sampling theory. 
Second, methods that depend on inducing points reach a plateau when $\Delta/\xi_r$ is relatively large (Fig.~\ref{fig:simu_err_time_p}(a)) and thus are limited in data efficiency. 
Third, when $\Delta/\xi_r$ becomes sufficiently small (Fig.~\ref{fig:simu_err_time_p}(b)), this limitation no longer exists.
Based on these observations, we recommend using exact GP when both $\Delta/\xi_r$ and $n_r$ are sufficiently large, FITC-GP or KISS-GP when $\Delta/\xi_r$ is sufficiently small and $n_r$ is sufficiently large, and LGSWD-GP when $\Delta/\xi_r$ is sufficient small and $n_r$ is large.
In other words, among the methods shown, our method is the most accurate and time-efficient one in the regime where the latent grid is relatively fine and the number of training points is large. 



\section{Time Complexity Analysis}\label{sec:time}

For $N$ training data points on the grid, \eqref{GP_mean} and \eqref{eq:swd_inverse} gives the predictive mean for SWD-GP:  
\begin{equation}
\mu_* = \sum_{i=1}^{N}\frac{1}{\lambda_i}({\bf k}^T_* {\rm v}_i) ({\rm v}_i^T {\bf y}).
\label{eq:swd_mean}
\end{equation}
Since the second inner product has time complexity $O(N)$, the sum has a total time complexity of $O(N^2)$. The space complexity of SWD-GP is $O(N)$. For $M$ inducing points on a grid and $N$ training data points, from Sec.~\ref{sec:latent} one can show that the predictive mean for LGSWD-GP is 
\begin{equation}
\mu_* = \sum_{i=1}^{M}\frac{1}{\chi_i}(k_{\bf *g} {\rm u}_i) ({\rm u}_i^T K_{\bf gx}\Lambda^{-1}{\bf y}),
\label{eq:lgswd_mean}
\end{equation}
where $\chi_i = \lambda_i + {\rm u}_i^T K_{\bf gx} \Lambda^{-1} K_{\bf xg} {\rm u}_i$.
The time complexity 
for 
computing $\chi_i$ is $O(pN)$, where $p$ is the number of bands used in the approximation ($p=3$ for tridiagonal and $p=5$ for pentadiagonal); thus, the sum gives a total time complexity of $O(MpN)$.
The same time complexity is found for computing $\Lambda^{-1}$ and $Q^{-1}$.
The space complexity of LGSWD-GP is $O(pN)$ with the bottleneck being at representing $K_{\bf gx}$.


Next, for clarity of comparison, we show how our method differs from other methods in terms of the other methods' bottleneck operation: Exact GP requires composing the kernel matrix and inverting it. The bottleneck operation is the inverse, which has time complexity $O(N^3)$ for $N$ training points. If the data lie on a grid, the SWD method can compose the inverted kernel analytically via \eqref{eq:swd_inverse}, although in practice, as shown by \eqref{eq:swd_mean}, we never need to explicitly represent the inverted kernel matrix for computing the predictive mean.

The bottleneck operation of FITC-GP is the matrix multiplication $K_{\bf gx}\Lambda^{-1}K_{\bf xg}$ in \eqref{matrix_Q}, which has time complexity $O(M^2N)$ for $M$ inducing points and $N$ training points. If the inducing points lie on a grid, that multiplication in LGSWD-GP with $p$-diagonal kernel approximation has time complexity $O(MpN)$, because each row in $K_{\bf xg}$ has at most $p$ non-zero elements. 
Note that one can choose not to explicitly compute this particular matrix product as it is absorbed into the $\chi_i$'s in \eqref{eq:lgswd_mean}. 
In addition, whenever a matrix inverse operation is required, such as in \eqref{eq:sigma_x}, LGSWD-GP benefits from being able to construct the inverse directly.

KISS-GP, which also uses inducing points on a grid, replaces all the operations in computing the predictive mean in \eqref{general_data_prediction} by solving a linear system $\Tilde{K}_{\bf *x}\Tilde{K}_{\bf xx}^{-1}{\bf y}$ through linear conjugate gradients, where $\Tilde{K}_{\bf ab} \approx K_{\bf ag} K_{\bf gg}^{-1} K_{\bf gb}$ \cite{wilson2015kernel}. The optimization has time complexity $O(SqN)$ because the product $\Tilde{K}_{\bf xx}{\bf y}$ can be computed in $O(qN)$ given the sparse approximation made by KISS-GP and because the optimization can be fixed to $S$ steps. 
The contribution of $q$ and $k$ in KISS-GP correspond to the contribution of $p$ and $M$ in LGSWD-GP, respectively: $p$ and $q$ stem from sparse approximations, while $M$ and a good choice for $S$ depend on the number of contributing eigenvalues. 


Lastly, we mention that there are a couple of ways to further speed up SWD-GP and LGSWD-GP. First, as a result of the sine functions appeared in $\{{\rm v}_k\}_{k=1}^N$, we may rewrite the mean and variance in (\ref{GP_mean}) and (\ref{GP_variance}) in terms of Discrete Fourier Transformed vector for the training data $\{y_i\}_{i=1}^N$, 
\begin{equation}\label{eq:dft}
    [{\rm FFT}_{\theta}({\bf y})]_j = \sum_{k=1}^N e^{ik\theta_j}y_k\:,
\end{equation} where $\theta_j = j\pi/(N+1)$. As such, the mean function becomes
\begin{equation}
    \mu_* = \frac{2}{N+1}\sum_{i=1}^N\frac{{\mathcal Im}[{\rm FFT}_{\theta}(k_*)]_i\ {\mathcal Im}[{\rm FFT}_{\theta}({\bf y})]_i}
    {\lambda_i}\:, 
\end{equation} and the predicting variance is
\begin{equation}
    \sigma^2_* = \sigma^2 - \frac{2}{N+1}\sum_{i=1}^N\frac{\left\{{\mathcal Im}[{\rm FFT}_{\theta}(k_*)]_i\right\}^2}
    {\lambda_i}\:.
\end{equation}
For SWD-GP, this reduces the time complexity from $O(N^2)$ to $O(N\log N)$, analogous to how Toeplitz structure reduces the time complexity of matrix multiplication. 
For LGSWD-GP, although this will not help reduce the bottleneck time complexity, it can help reduce many of the operations involving matrix products with the eigenvectors.
Second, the sum in the eigen-decomposition of \eqref{EigenBasis} and \eqref{eq:eigencomp} can be computed independently and is directly amenable to parallelization.

\section{Related Work}\label{sec:relatedW}

Previous works have also taken advantage of structured data to increase the computational efficiency of GP.
For input data that lie on a grid, \cite{cunningham2008fast,turner2010statistical} sidestep the matrix inversion by solving $K^{-1}y$ with gradient-based methods. Importantly, the optimization is fast because Toeplitz / circulant structure can be exploited for fast matrix multiplication. These fast multiplications take on forms similar to \eqref{eq:dft}.
For input data that do not lie on a grid, \cite{wilson2015kernel} (KISS-GP) introduces a latent grid, finds a sparse, approximate representation of $K_{\bf xg}$ in terms of $K_{\bf gg}$, and makes inference using the same kind of optimization problem in \cite{cunningham2008fast,turner2010statistical}. The sparsification in KISS-GP and that in our LGSWD-GP share the same flavor in ignoring the diminishing long-range correlations among data points.
Recently, \cite{evans2017scalable} further extended this optimization approach to efficiently handle high dimensional data by applying the Nystr\"{o}m approximation for eigen-decomposition and by exploiting properties of Kronecker and Khatri-Rao products for fast matrix multiplication. 
Overall, our method and these methods all take advantage of the grid structure to form fast matrix multiplication; however, our method differs in that it overcomes the matrix inversion bottleneck via analytic diagonalizaion rather than optimization.   


Variational approaches are another popular method for GP inference. Variational methods turn the inference of the predictive mean and variance into an optimization problem \cite{titsias2009variational}. Variational GP often provides more accurate prediction than FITC \cite{bauer2016understanding} and has been made progressively faster through a series of development---from stochastic variational inference \cite{hensman2013gaussian}, to distributed variational inference \cite{gal2014distributed}, to asynchronous distributed variational inference \cite{peng2017asynchronous}---to handle billions of input data.
Similar to our work and \cite{evans2017scalable}, \cite{nickson2015blitzkriging} uses grid inducing points with stochastic variational inference, which allows added efficiencies in computation via the Kronecker and Khatri-Rao products.  


\section{Discussion}\label{sec:diss}

In this paper we present an analytic approach to invert a class of kernels for GP regression with grid inputs through standing wave decomposition (SWD). This class of kernels is roughly Toeplitz \footnote{The tridiagonal kernel is exactly Toeplitz. For pentadiagonal kernel, \eqref{eq:kij_mod} shows which elements are modified and by how much. The procedure is readily extendable to more diagonials.} and thus approximates all translational invariant kernels with grid inputs. Here we show results for tridiagonal Toeplitz kernels, which can be thought of as approximating a square exponential kernel by keeping only the nearest neighbor correlation among data points, and a pentadiagonal kernel, which includes the next nearest neighbor correlation. We demonstrate that the approach can be extended to approximate product kernels in higher dimensions. Lastly, we apply the approach to a latent grid of inducing points so as to handle training data that do not lie on a grid. 

From the perspective of signal processing, one may regard the grid as a collection of input points at which the function is sampled. Thus, the grid representation of data and function is sufficient for prediction and reconstruction as long as the underlying functions have a finite Nyquist rate.

Our SWD approach is unique in that the inversion of the (approximated) kernel matrix is done analytically. This analytic inverse, along with the sparse matrices induced by keeping only the most important correlations among data points, makes this approach computationally very efficient (Sec.~\ref{sec:time}). Furthermore, the optimal length scale hyperparameter in the SWD approach can be determined without the need to invoke optimization (Sec.~\ref{sec:recons}).
When applied to training data on a grid, we see that the predictions of SWD-GP are very similar to those of exact GP if the two use the same length scale (Fig.~\ref{RegressionReconstructed}), although vastly faster.
For data that do not lie on a grid, our results show that LGSWD-GP can be both faster and more accurate than existing approximation GP methods when the number of training points is large (Figs.~\ref{fig:simu_err_time} and \ref{fig:simu_err_time_p}).
We expect that this advantage will be magnified in situations where a highly precise latent representation is not crucial, such as GP classification.
The analytical form of the eigen-decomposition also allows straight-forward parallelization of our approach, which we leave to future work. Other future directions include extending the approach to kernel approximations with longer-range correlation, to other types kernel such as periodic kernels, and to latent grid in high dimensions.



\acknowledgements
This material is based on research sponsored by the Air Force research Laboratory and DARPA under agreement number FA8750-17-2-0146 to P.S. and S.Y. The U.S. Government is authorized to reproduce and distribute reprints for Governmental purposes notwithstanding any copyright notation thereon.

\bibliographystyle{unsrt}
\end{document}